\documentclass[conference]{IEEEtran}
\IEEEoverridecommandlockouts

\usepackage{tabularx}
\usepackage[table]{xcolor}
\usepackage{url}
\usepackage{cite}
\usepackage{amsmath,amssymb,amsfonts}
\usepackage{algorithmic}
\usepackage{graphicx}
\usepackage{textcomp}
\usepackage{xcolor}
\usepackage{mdframed}
\usepackage{enumitem}

\def\BibTeX{{\rm B\kern-.05em{\sc i\kern-.025em b}\kern-.08em
    T\kern-.1667em\lower.7ex\hbox{E}\kern-.125emX}}
\begin{document}

\title{Agentic Service-Oriented Computing:\\A Manifesto for the Next Frontier of Service-Oriented Computing
}

% \author{Amin Beheshti, Rong Chang, et al.
% }

\author{
\IEEEauthorblockN{
Amin Beheshti\IEEEauthorrefmark{1},
Rong N. Chang\IEEEauthorrefmark{2},
Boualem Benatallah\IEEEauthorrefmark{3},
Fabio Casati\IEEEauthorrefmark{4},
Schahram Dustdar\IEEEauthorrefmark{5},\\
Geoffrey Fox\IEEEauthorrefmark{6},
Quan Z. Sheng\IEEEauthorrefmark{1},
Yan Wang\IEEEauthorrefmark{1},
Jian Yang\IEEEauthorrefmark{7},
Albert Zomaya\IEEEauthorrefmark{8}
}
\IEEEauthorblockA{\IEEEauthorrefmark{1}Macquarie University, Sydney, Australia}
\IEEEauthorblockA{\IEEEauthorrefmark{2}IBM Research, Yorktown Heights, USA}
\IEEEauthorblockA{\IEEEauthorrefmark{3}School of Computing, Dublin City University, Dublin, Ireland}
\IEEEauthorblockA{\IEEEauthorrefmark{4}University in Trento, Trento, Italy}
\IEEEauthorblockA{\IEEEauthorrefmark{5}TU Wien, Vienna, Austria}
\IEEEauthorblockA{\IEEEauthorrefmark{6}University of Virginia, Charlottesville, USA}
\IEEEauthorblockA{\IEEEauthorrefmark{7}Beijing Normal--Hong Kong Baptist University (BNBU), Zhuhai, China}
\IEEEauthorblockA{\IEEEauthorrefmark{8}The University of Sydney, Sydney, Australia}
}

\maketitle

\begin{abstract}
The rapid emergence of LLM-powered autonomous and semi-autonomous agents is reshaping software systems from static, request-response components into goal-directed, adaptive, and tool-using computational actors. As these agents move from isolated cognitive prototypes into complex distributed workflows, they confront challenges that the Service-Oriented Computing community has studied for more than two decades: composition, interoperability, quality of service, lifecycle management, governance, security, and trust. Yet much of today's agentic AI ecosystem is developing these foundations 
ad hoc, without the engineering rigour required for dependable enterprise and societal deployment. This paper introduces Agentic Service-Oriented Computing (ASOC) as a new research and practice area concerned with engineering agents as services, orchestrating services through autonomous and semi-autonomous agents, and governing ecosystems of agents and services under constraints of trust, cybersecurity, compliance, performance, and accountability. We articulate six foundational principles of ASOC (harness-ability, composability, lifecycle engineering, trustworthiness by design, goal-driven orchestration, and observability/accountability) and organise a five-dimensional research agenda spanning: (i) agentic services foundations and lifecycle engineering; (ii) composition, orchestration, and interoperability; (iii) governance, observability, and accountability; (iv) security, trust, and risk management; and (v) evaluation, certification, and Agentic QoS. We argue that the Services Computing community is especially well positioned to provide the conceptual and engineering spine for this emerging field, transforming agentic AI from fragmented demonstrations into dependable, service-based systems worthy of human and organisational trust.
\end{abstract}


\section{The Inflection Point}

Software systems are undergoing a transformation that is qualitatively different from prior generational shifts. Each preceding paradigm (from spaghetti code and structured programming through object-oriented design to the interoperability and composability achieved by Service-Oriented Computing and microservices architectures) addressed the limitations of its predecessor by introducing a new level of abstraction, modularity, and reusability~\cite{ref18,ref22,ref26}. Yet each of these transitions preserved a fundamental constant: the human developer remained the author of explicit computational logic, translating intent into instructions that machines would execute. Natural language is now becoming a high-level medium for specifying goals, constraints, and desired outcomes, while agents increasingly translate those specifications into executable plans, service calls, and adaptive workflows~\cite{ref18}. This shift is not merely a technical upgrade; it changes the nature of the software artefact itself.

LLM-powered agents are not the first autonomous software entities. Autonomous agents, Belief–desire–intention (BDI) architectures, multi-agent systems, workflow agents, and adaptive systems have been studied for decades~\cite{ref23}. What distinguishes the current generation is something more specific: LLM-powered agents are the first class of broadly deployable, natural-language-mediated software entities that combine reasoning, planning, tool use, service invocation, memory, and adaptive execution in a form accessible to mainstream enterprise and societal deployment. Unlike conventional service artefacts, which are typically invoked with a fully specified request and expected to return a bounded response, an LLM-powered agent may receive an under-specified goal and determine how to pursue it. It perceives context, engages in context-sensitive decision-making under incomplete specifications, selects and sequences actions, delegates subtasks to other agents or services, recovers from failures, and adapts dynamically to environmental feedback.

This shift is already visible in practice. LLM-powered agents are being deployed across enterprise workflows, scientific discovery pipelines, customer service systems, software engineering processes, and public-sector services. Ecosystem platforms such as LangGraph\footnote{\url{https://www.ibm.com/think/topics/langgraph}},
CrewAI\footnote{\url{https://crewai.com/}}, 
AutoGen\footnote{\url{https://microsoft.github.io/autogen/stable//index.html}}, 
Microsoft Copilot Studio\footnote{\url{https://www.microsoft.com/en-us/microsoft-365-copilot/}}, 
OpenAI's Codex\footnote{\url{https://chatgpt.com/codex/}} (Responses API/Agents SDK/AgentKit), and related agent frameworks provide scaffolding for building and deploying agentic workflows; these platforms are illustrative of a fast-moving ecosystem rather than an exhaustive inventory. Model Context Protocol (MCP)~\cite{ref24} has emerged as an open protocol for connecting LLM applications with tools and external data sources, while Agent2Agent (A2A)~\cite{ref29} was introduced to support communication, secure information exchange, and coordination among agents across enterprise platforms. MCP and A2A are important early signs that the agent ecosystem is moving toward interoperability, but, by themselves, they do not solve the full Agentic Service-Oriented Computing (ASOC) problem. ASOC must define the surrounding engineering framework: capability contracts, delegation semantics, lifecycle management, Agentic QoS, trust calibration, compliance evidence, and harness-based governance. They are scaffolds, not foundations.

However, beneath the impressive surface of agent capabilities lies a profound and growing engineering deficit. Current agentic systems are overwhelmingly ad hoc in design, weakly governed in operation, opaque in decision-making, and poorly equipped to meet the security, compliance, and reliability demands of enterprise and societal deployment. Agents are typically built as isolated cognitive prototypes. They lack version-controlled interfaces, enforced service-level agreements, runtime monitoring, systematic failure recovery, and the interoperability contracts that make distributed systems dependable. The field is, in effect, repeating the mistakes of pre-SOA distributed computing: \emph{impressive capabilities assembled without the engineering discipline needed to make them trustworthy at scale.}

This is precisely the moment at which the service-oriented computing community must act. The questions raised by agentic AI (\emph{i.e., How agents are designed, composed, orchestrated, governed, monitored, and trusted in real-world environments}) are, at their core, service-oriented computing questions. They are the same questions our community has been studying and answering since the foundational works of the early 2000s~\cite{ref1,ref2,ref22}, now instantiated at a new level of autonomy and intelligence. The transition from Web services, service ecosystems, and microservices to agentic services is not a departure from our intellectual tradition; it is its most ambitious expression.

This paper is both a scholarly argument and a call to action. This manifesto makes four contributions:

\begin{itemize}
    \item First, it defines Agentic Service-Oriented Computing (ASOC) as an emerging research and practice area grounded in, but extending beyond, Service-Oriented Computing (SOC). 
    \item Second, it introduces a tripartite framing of ASOC through agents-as-services, services orchestrated by agents, and agent-service ecosystems governed. 
    \item Third, it articulates six foundational engineering principles for dependable agentic services.
    \item Fourth, it proposes a five-dimensional research agenda covering lifecycle engineering, composition, orchestration, governance, cybersecurity, trust, and evaluation.
\end{itemize}

The remainder of this paper is organised as follows. 
Section~II surveys the heritage of Service-Oriented Computing and identifies the seven foundational pillars that ASOC inherits and transforms. 
Section~III examines the rise of LLM-powered agentic AI, characterising its cognitive capabilities and illustrating the stakes through four domain-specific use cases. 
Section~IV formally defines Agentic Services Computing, introduces the four core technical concepts (agentic service, delegation contract, agent harness, and agentic service ecosystem), presents the ASOC reference architecture, and positions ASOC relative to adjacent research areas. 
Section~V articulates the six foundational engineering principles of ASOC, each stated with a definition, motivation, core research questions, and concrete realisation mechanisms. 
Section~VI proposes the five-dimensional ASOC research agenda spanning agentic service foundations and lifecycle engineering; composition, orchestration, and interoperability; governance, observability, and accountability; security, trust, and risk management; and evaluation, certification, and Agentic QoS. 
Section~VII makes the case for the Services Computing community leadership of this field. 
Section~VIII concludes with a call to action for the research and practitioner community.

\section{The Heritage of Service-Oriented Computing (SOC)}

The Services Computing community has spent more than two decades solving precisely the engineering problems that agentic AI now faces. This section articulates what ASOC inherits. Seven foundational concepts of service-oriented computing carry directly into the agentic context, each becoming harder to realise but no less necessary.

\subsection{Seven Pillars of SOC that ASOC Inherits}

\begin{table*}[t]
\centering
\caption{From SOC to ASOC: Transformation of foundational Service-Oriented Computing\\pillars for probabilistic, goal-directed agentic services.}
\label{table1}
\renewcommand{\arraystretch}{1.25}
\setlength{\tabcolsep}{8pt}
\begin{tabularx}{\textwidth}{|p{0.27\textwidth}|X|}
\hline
\rowcolor[HTML]{D9EAF7}
\textbf{SOC Inheritance} & \textbf{ASOC Transformation} \\
\hline

\textbf{Service contract} &
Agentic capability and delegation contract (goal types, uncertainty, trust, revocability) \\
\hline

\textbf{Publish-find-bind} &
Goal-level agentic service discovery via capability and intent matching \\
\hline

\textbf{Service composition} &
Goal-driven orchestration: agent-assisted, runtime service composition under formal constraints \\
\hline

\textbf{QoS / SLA} &
Agentic QoS integrating latency, factual accuracy, goal fidelity, compliance, and trust-calibrated SLAs \\
\hline

\textbf{Lifecycle engineering} &
Agent Development Lifecycle (ADLC) incorporating drift detection, model versioning, certification \\
\hline

\textbf{Monitoring and auditability} &
Agent provenance graphs and semantic audit logs spanning the full reasoning-to-action chain \\
\hline

\textbf{Governance} &
Harness-based governance and policy enforcement: delegation scope, tool permissions, approval gates, revocation, compliance evidence \\
\hline

\end{tabularx}
\end{table*}

%%%%%%%%%%%%%%%%%

\textbf{Service contracts and interfaces.} Papazoglou and Georgakopoulos established that every SOC interaction rests on a well-defined, published service contract~\cite{ref1}. Papazoglou's comprehensive textbook~\cite{ref22} remains the definitive treatment of these principles. The Extended Service-Oriented Architecture~\cite{ref2} introduced the layered pyramid of Basic, Composite, and Managed Services, establishing that a mature service infrastructure is a governed ecosystem. In ASOC, service contracts must be extended to describe goal types, uncertainty characteristics, execution constraints, and trust requirements.

\textbf{Discovery and binding.} The publish-find-bind model~\cite{ref1} established that services must be discoverable by principals who did not participate in their design. In ASOC, discovery must work at the goal level: an orchestrating agent must find, assess, and bind to agentic services whose capabilities match a specified goal, not merely a typed interface.

\textbf{Composition and orchestration.} Service composition was the defining research challenge of early SOC. Casati et al.'s eFlow~\cite{ref3} demonstrated adaptive, runtime-modifiable composition. Self-SERV~\cite{ref4} introduced declarative, peer-to-peer provisioning. BPEL4WS~\cite{ref5} provided an executable process notation. McIlraith and Son brought AI planning to semantic composition~\cite{ref6}. Surveys by Dustdar and Schreiner~\cite{ref28} and Lemos, Daniel, and Benatallah~\cite{ref13} map this landscape from manual to fully automated approaches. Context-aware specification techniques such as ContextUML~\cite{ref21} anticipate the richer capability descriptions agentic services require. In ASOC, composition must operate at the goal level, with an intelligent agent rather than a static workflow engine as the composer.

\textbf{Quality of Service and SLAs.} Zeng et al.~\cite{ref7} demonstrated that service selection must optimise composite non-functional properties including latency, reliability, cost, and availability. This insight (that engineering must account for emergent QoS properties of composed systems) remains one of the SOC community's most important contributions and is almost entirely absent from current agentic AI engineering.

\textbf{Lifecycle engineering.} The SODDM~\cite{ref20} established a formal service lifecycle: analysis, design, implementation, testing, deployment, management, and evolution. In ASOC, this lifecycle must be extended to accommodate probabilistic, goal-directed agents whose behaviour is influenced by underlying models.

\textbf{Governance and trust.} Papazoglou and van den Heuvel showed that governance is not optional in service systems~\cite{ref10}. The SOC Research Roadmap of 2008~\cite{ref8} formalised this across four themes: foundations, composition, management and monitoring, and engineering.

\textbf{Monitoring, observability, and auditability.} Work on deriving protocol models from service interaction logs~\cite{ref27} established the principle that service behaviour can and must be observed and analysed at runtime. In ASOC, where agent reasoning is probabilistic and internally opaque, observability and auditability are even more critical and technically harder to achieve. Table~\ref{table1} summarises how each of the seven SOC pillars is transformed in the ASOC context. For each inherited concept, the table shows what the SOC community built, and what ASOC must now extend, harden, or reconstitute to handle probabilistic, goal-directed agentic behaviour.

\subsection{From Web Services to Cloud and Microservices}

The SOC tradition evolved steadily in response to technological change. Fielding's REST~\cite{ref9} enabled the API economy upon which LLM tool-calling now operates. The microservices movement~\cite{ref11} decomposed the service monolith into independently deployable bounded services, anticipating cognitive capability at the service boundary. The NIST cloud model~\cite{ref12} completed the \emph{as-a-service} abstraction. The intellectual heritage of agent-based software engineering~\cite{ref23} constitutes an important precursor, particularly regarding the design of autonomous, goal-directed entities in dynamic environments. Across twenty-five years of research, the SOC community built and validated engineering principles (contracts, discoverability, composability, QoS, lifecycle, governance, and observability) that now await reinstatement in the agentic context.

\section{The Rise of Agentic AI: Promise and Peril}

\subsection{LLM-Powered Agents: What They Can Do}

The emergence of Large Language Models (LLMs) as cognitive cores for software agents represents a fundamental inflection in capabilities. LLM-based agents combine natural language understanding with reasoning, planning, and tool invocation, enabling systems that can pursue complex, multi-step objectives with minimal procedural specification~\cite{ref14,ref26}. Chain-of-thought prompting~\cite{ref19} demonstrated that eliciting intermediate reasoning dramatically improves performance on complex tasks. The ReAct framework~\cite{ref15} extended this by interleaving reasoning traces with action execution, establishing the sense-reason-act cycle as the fundamental computational unit of agentic behaviour. Executable code actions~\cite{ref25} further extended agent capabilities, enabling reliable tool use across complex domains.

Early work such as ProcessGPT~\cite{procesGPT} demonstrated how generative AI can transform business process management by replacing static process models with LLM-driven process generation, adaptation, and automation; an important forerunner of the goal-driven orchestration paradigm central to ASOC.
Multi-agent frameworks have extended these capabilities into collaborative, division-of-labour systems. MetaGPT~\cite{ref16} demonstrated that assigning LLM agents to structured organisational roles with shared memory and standard operating procedures significantly improves outcomes. AutoGen, CrewAI, and LangGraph provide general frameworks for building multi-agent pipelines. Wang et al.'s survey~\cite{ref14} established the canonical taxonomy of agent capabilities: memory, planning (including chain-of-thought~\cite{ref19}), tool use (including code execution~\cite{ref25}), and environment interaction. These represent a qualitative expansion of what software agents can do; but they are cognitive capabilities, not engineering ones. These frameworks are important early scaffolds. They do not yet constitute a mature service-engineering discipline.

\subsection{The Engineering Deficit}

Despite impressive cognitive capabilities, the engineering discipline surrounding LLM-based agents remains strikingly immature. The closest related work is Deng et al.~\cite{ref17}, introducing Agentic Services Computing (ASC) as a paradigm organised around a four-phase lifecycle (Design, Deployment, Operation, Evolution) and four research dimensions: perception and context modelling, autonomous decision-making, multi-agent collaboration, and evaluation with alignment and trustworthiness. That work makes substantial contributions: it provides a formal definition of ASC, introduces the SCALE framework characterising five properties of agentic services, proposes Agent Service Level Objectives (ASLOs), and surveys more than 200 representative works mapping research advances to each lifecycle phase and dimension. It is, at its core, a comprehensive survey and paradigmatic framework paper. 

Our contribution is complementary: whereas Deng et al.~\cite{ref17} provide a broad survey and lifecycle-oriented framework, this manifesto develops a normative service-engineering doctrine, reference architecture, and community agenda for ASOC.
In particular, this manifesto presents: six normative engineering principles, each with realisation mechanisms; an ASOC reference architecture as a reusable community artefact; specific technical concepts of delegation contract, agent harness, and agentic service ecosystem as defined engineering objects; a structured SOC-to-ASOC inheritance argument grounded in seven pillars; an ASOC-specific threat model; and an explicit strategic case for Services Computing community leadership. 
The present manifesto is complementary but distinct in purpose, scope, and contribution. Where Deng et al.~\cite{ref17} ask \textit{what has been done and how should agentic services be structured and evaluated across a lifecycle?}, this manifesto asks a prior and normative question: \textit{what engineering doctrine must govern this field, and how should the Services Computing community organise to lead it?} Table~\ref{tab:comparison} provides a concrete side-by-side comparison.

Conversely, this manifesto does not attempt the comprehensive literature survey that Deng et al. achieve, nor does it map technical research to lifecycle phases at that level of granularity. Together, the two works are complementary: Deng et al.~\cite{ref17} provide the paradigmatic survey foundation; this manifesto provides the normative service-engineering doctrine, reference architecture, and community agenda required to consolidate ASOC as a research field with a distinct identity and leadership programme.

\begin{table*}[t]
\caption{Side-by-side comparison: Deng et al.~\cite{ref17} versus this manifesto}
\label{tab:comparison}
\centering
\renewcommand{\arraystretch}{1.3}
\begin{tabularx}{\textwidth}{>{\bfseries}p{0.13\textwidth} p{0.40\textwidth} p{0.40\textwidth}}
\hline
\rowcolor[HTML]{D9EAF7}
\textbf{Dimension} & \textbf{Deng et al.~\cite{ref17}} & \textbf{This manifesto} \\
\hline
Nature of paper &
  Comprehensive survey (200+ works) and paradigmatic framework &
  Normative position/manifesto paper with prescriptive engineering doctrine \\

Primary question &
  What has been done, and how should agentic services be structured, evaluated, and governed across a lifecycle? &
  What engineering doctrine must govern ASOC as a field, and how should the Services Computing community lead it? \\

Field definition &
  Formal Definition 1 provided; lifecycle- and capability-driven paradigm &
  Formal definition emphasising tripartite scope: agents-as-services, agent-driven orchestration, governed ecosystems \\

Research framework &
  Two-dimensional matrix: four lifecycle phases (horizontal) $\times$ four research dimensions (vertical): perception, decision-making, collaboration, evaluation/trustworthiness &
  Tripartite framing plus five normative research dimensions with open research questions \\

Core concepts introduced &
  SCALE framework (five agentic service characteristics); ASLOs (Agent Service Level Objectives) &
  Delegation contract, agent harness, agentic service ecosystem, Agentic QoS, agent provenance graph \\

Governance content &
  Agent registries, retirement protocols, audit APIs, ASLOs, Constitutional AI, reputation systems, DAOs &
  Harness-based governance; six principles, each with realisation mechanisms; ASOC-specific threat model \\

Engineering principles &
  None stated normatively &
  Six normative principles, each with definition, motivation, research questions, and realisation mechanisms \\

Reference architecture &
  None &
  ASOC six-layer stack with Governance/Observability and Security/Trust cross-cutting planes \\

SOC grounding &
  Section II covering SOA, microservices, serverless, BPEL, SLAs, monitoring tools &
  Seven inherited SOC pillars with explicit SOC-to-ASOC transformation table; extension of the SOC Research Roadmap \\

Literature coverage &
  Comprehensive survey of 200+ works across all dimensions &
  Selective citations grounding principles, and agenda; not a survey \\

Community agenda &
  Three-phase milestones roadmap; calls for registries, ASLOs, and interoperability standards &
  Explicit strategic case for Services Computing community leadership; call for standards, benchmarks, governance frameworks \\
\hline
\end{tabularx}
\end{table*}

Taken together, LLM-based agents are frequently developed as isolated cognitive prototypes, lacking essential engineering features such as version control, runtime monitoring, SLA enforcement, and systematic failure recovery. This structural deficiency makes current agentic systems unsuitable for enterprise and public-sector deployment.
Five specific deficits deserve emphasis:

\begin{itemize}
    \item \emph{Agentic systems lack formal capability contracts:} unlike WSDL-described web services~\cite{WSDL} or OpenAPI-specified\footnote{https://swagger.io/specification/} REST APIs, agents advertise capabilities informally through unverified, non-binding natural language descriptions.
    \item \emph{Agent composition is predominantly ad hoc:} multi-agent workflows are assembled through programmatic scaffolding rather than declarative composition models supporting verification, optimisation, and runtime adaptation.
    \item \emph{QoS monitoring is largely absent:} no established frameworks measure agent response quality (latency, reliability, factual accuracy, policy compliance) across composed agent workflows.
    \item \emph{Agent lifecycle management is nascent:} no established practices exist for versioning agents, managing deployment, detecting behavioural drift, or governing evolution after model updates.
    \item \emph{Security engineering is critically underdeveloped:} agents are vulnerable to prompt injection, data exfiltration, privilege escalation, and supply chain attacks via malicious tool definitions, operating at a layer below existing security frameworks.
\end{itemize}

\subsection{The Fragmentation Risk}

The most consequential systemic risk facing agentic AI is fragmentation: the proliferation of incompatible frameworks, proprietary protocols, and isolated capability silos lacking the interoperability, portability, and governance scaffolding needed for enterprise and societal deployment. This is the precise trajectory observed in distributed computing before Service Oriented Architecture (SOA)~\cite{ref10}, in cloud computing before NIST's (National Institute of Standards and Technology) reference architecture\footnote{\url{https://csrc.nist.gov/pubs/sp/800/145/final}}, and in earlier AI-agent research before fragmentation into incompatible platforms. Without strong disciplinary leadership grounded in rigorous service engineering, the agentic AI field will recapitulate these cycles at accelerated speed.

\subsection{Illustrative Use Cases and Stakes}

The stakes of ASOC are clearest where agents are delegated 
consequential goals rather than narrow tasks. The central 
question is not only whether an agent can complete a task, 
but whether its actions are contractually bounded, observable, 
governable, and accountable to human and organisational authority.

\emph{Healthcare Agentic Service:} 
A clinical workflow agent coordinating patient records, referral 
pathways, and clinician approvals must operate under a strong 
engineering regime: a delegation contract bounding its authorised 
scope, a harness enforcing human approval before consequential 
actions, and an audit trail recording evidence sources, tool calls, 
and recommendation pathways. Capable agents are not deployable 
services until autonomy is bounded by contracts, governance 
metadata, and observable execution structures.

\emph{Enterprise Procurement Agent:} 
An agent delegated to identify suppliers, compare quotes, and 
prepare purchase orders may act within technical permissions while 
exceeding the organisation's intended delegation scope. An agentic 
service therefore requires explicit financial authority boundaries, 
approval gates, revocation mechanisms, and immutable transaction 
logs, making delegation and harness-based governance first-class 
engineering concerns.

\emph{Scientific Discovery Pipeline:} 
Where multiple specialised agents search literature, invoke 
simulations, and compose workflows, value depends not only on 
reasoning ability but on reproducibility and scientific 
defensibility. ASOC provides the engineering frame: capability 
contracts clarify contributions; Agentic QoS captures reliability 
and uncertainty; provenance~\cite{provenance} graphs record 
intermediate artefacts, service calls, and decision points; and 
governance mechanisms manage error propagation across the pipeline.

\emph{Public-Sector Citizen Service Agent:} 
An agent helping citizens navigate benefits systems must address 
fairness, explainability, and compliance with statutory obligations, 
not merely task automation. It must distinguish advice from 
determination, maintain records of policy sources and decision 
pathways, support human appeal, and monitor for systematic 
performance differences across population groups.

Across these examples, provenance is necessary but not sufficient. 
What makes each system an agentic service is the combination of 
provenance with service contracts, delegation semantics, runtime 
harnessing, policy enforcement, security controls, Agentic QoS, and lifecycle governance. These examples substantiate the manifesto's central claim: the future of agentic AI depends on a service-engineering discipline capable of making agents dependable, composable, governable, observable, secure, and worthy of trust.

\section{Defining Agentic Service-Oriented Computing (ASOC)}

\subsection{ A Formal Definition}

We offer the following \textbf{definition} of the new field:

\begin{mdframed}[linewidth=1pt, innerleftmargin=8pt, innerrightmargin=8pt, innertopmargin=6pt, innerbottommargin=6pt]
\emph{\textbf{Agentic Service-Oriented Computing (ASOC)} is the research and practice area concerned with the design, engineering, deployment, orchestration, governance, and lifecycle management of AI-driven services and service ecosystems in which autonomous or semi-autonomous agents reason, plan, invoke tools, coordinate with other agents, and act on behalf of humans or organisations under explicit constraints of trust, cybersecurity, compliance, performance, accountability, and societal value.}
\end{mdframed}

\vspace{6pt}
\noindent
This definition encompasses three nested concerns:

\begin{enumerate}[label=\roman*.]    
\item \emph{Agents as Services:} engineering AI agents as harnessable, composable, and governable service entities with explicit capability contracts, lifecycle semantics, and QoS.
    \item \emph{Services Orchestrated by Agents:} orchestration of distributed services by autonomous/semi-autonomous agents operating under human oversight and policy constraints.
    \item \emph{Governed Agent-Service Ecosystems:} governance of heterogeneous agent-service populations under constraints of cybersecurity, trust, compliance, and societal value.
\end{enumerate}

\vspace{6pt}
\noindent
ASOC introduces four core technical concepts:

\begin{enumerate}[label=\roman*.]    
\item
\emph{Agentic Service:} A service entity that accepts goals, constraints, and context; evaluates possible actions; invokes tools or other services; and returns outcomes with provenance, confidence, cost, risk, and execution metadata.
\item
\emph{Delegation Contract:} A formal specification of the scope, authority, constraints, duration, revocability, and accountability conditions under which a human or organisation delegates a goal to an agentic service.
\item
\emph{Agent Harness:} A runtime governance wrapper that mediates an agent's goals, permissions, tools, memory, policies, execution traces, and escalation pathways. The agent harness is the core ASOC mechanism through which autonomy is converted into governed autonomy. It mediates the boundary between model capability and service accountability by enforcing delegation scope, tool permissions, policy constraints, execution logging, approval gates, circuit breakers, and revocation mechanisms.
\item
\emph{Agentic Service Ecosystem:} A governed, heterogeneous collection of agentic services and conventional services that can be discovered, composed, orchestrated, and managed under shared policies, trust frameworks, and interoperability standards.
\end{enumerate}

%%%%%%%%%%%%%%%%%%%%%%%%%%%%%
% Add to preamble if not already present:
% \usepackage{xcolor}
% \usepackage{colortbl}

\begin{table}[t]
\centering
\caption{Levels of Delegation and Autonomy in ASOC.
         Each level maps a degree of agent autonomy to
         the minimum ASOC control requirements a
         delegation contract must enforce.}
\label{tab:delegation-levels}
\small
\begin{tabular}{|p{0.55cm}|p{2.0cm}|p{4.6cm}|}
\hline
\rowcolor[HTML]{D9EAF7}
\textbf{Level} & \textbf{Description} & \textbf{ASOC Control Requirement} \\
\hline
L0 & Advisory only             & Provenance and explanation \\
\hline
L1 & Tool recommendation       & Human approval \\
\hline
L2 & Bounded execution         & Delegation contract and harness \\
\hline
L3 & Multi-step orchestration  & Continuous monitoring and revocation \\
\hline
L4 & High-consequence autonomy & Certification, audit, and liability model \\
\hline
\end{tabular}
\end{table}
%%%%%%%%%%%%%%%%%%%%%%%%%%%%%%%

Table~\ref{tab:delegation-levels} operationalises the delegation contract by mapping five levels of agent autonomy to their corresponding ASOC control requirements.
These definitions yield a concise characterisation of the core ASOC engineering object:

\begin{mdframed}[linewidth=1pt, innerleftmargin=8pt, innerrightmargin=8pt, innertopmargin=6pt, innerbottommargin=6pt]
\begin{center}    
\emph{Agentic Service  =  Agent  +  Service Contract  +\\Harness  +  Governance Metadata}
\end{center}
\end{mdframed}

In this formulation, the harness and governance metadata are what transform agent capability into service accountability. 

%\vspace{6pt}
\emph{An \textbf{Agent} becomes \textbf{Deployable} not when it can reason well, but when its reasoning is bounded by contracts, logged through a harness, and governed by explicit metadata that make its actions traceable and revocable.}

%\vspace{6pt}
%%% IMPORTANT
%Not all LLM agents satisfy the defining characteristics of an agentic service:

%\vspace{6pt}
\emph{An \textbf{Agent} becomes an \textbf{Agentic Service} only when it is engineered as a discoverable, composable, governable, observable, and accountable service entity with explicit capability, security, lifecycle, and QoS properties}. 

%\vspace{6pt}
The transition from agent to agentic service is precisely the engineering transformation that ASOC is concerned with.
%To make the field concrete, 
ASOC should produce a family of engineering artefacts that translate the manifesto's principles into implementable research objects: agentic capability contracts, delegation contracts, agent harnesses, agent provenance graphs, agentic service registries, Agentic QoS models, security profiles, compliance evidence schemas, and 
certification benchmarks.

\subsection{The ASOC Reference Architecture}

\begin{figure}[t]  
	\begin{center}
	\includegraphics[width=1.0\linewidth]{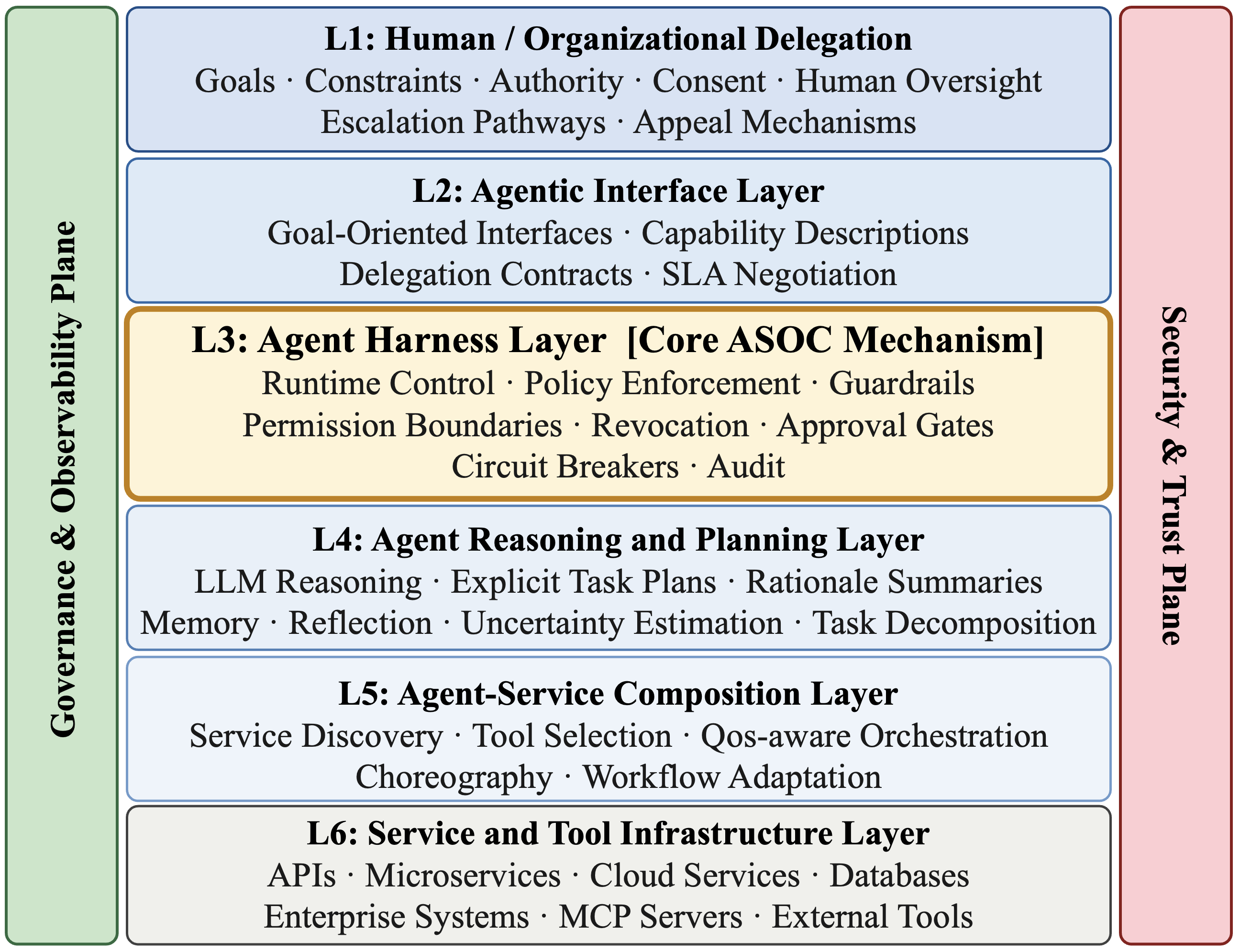}
	\end{center}
    	\caption{The Agentic Service-Oriented Computing Stack. Six functional layers from human/organisational delegation (L1) to service and tool infrastructure (L6), with Governance and Observability and Security and Trust as cross-cutting planes. The Agent Harness (L3, amber) is the core ASOC mechanism converting autonomy into governed autonomy.}
	\label{Fig1}
\end{figure}

Figure~\ref{Fig1} presents the Agentic Service-Oriented Computing Stack: the reference architecture that organises the technical concerns of ASOC. The stack comprises six functional layers (from human/organisational delegation at the top (L1), through the Agentic Interface (L2), the Agent Harness (L3, highlighted as the core ASOC mechanism), Agent Reasoning and Planning (L4), Agent-Service Composition (L5), and Service and Tool Infrastructure (L6) at the base) with two cross-cutting planes spanning all layers: Governance and Observability on the left, and Security and Trust on the right. These planes reflect that observability and security are not layer-specific concerns but pervasive engineering requirements of the entire stack.

\vspace{6pt}
\emph{
The Agent Harness (L3) is the flagship layer of the ASOC stack and the most important original concept introduced by this manifesto. Without a harness, an agent is a cognitive capability; with a harness, it becomes a deployable, governable agentic service. The agent harness is to Agentic Service-Oriented Computing what the service contract was to SOA, what the container runtime was to cloud-native microservices, and what the policy enforcement point is to governed enterprise systems.}

\subsection{Three Constitutive Lenses}

The agent-as-service lens re-frames the AI agent as a first-class service entity that exposes a goal-oriented interface. Capability descriptions for agentic services must draw on context-aware service modelling traditions~\cite{ref21} to support automated discovery, goal-level matching, and contractual accountability.

The service-orchestrated-by-agent lens 
Reframes the agent as an intelligent orchestrator that replaces a static workflow engine (such as BPEL~\cite{ref5}) with a dynamic, reasoning-capable orchestrator that, at runtime, determines which services to invoke and under what conditions based on a high-level goal specification. This raises fundamental questions about the safety, verifiability, and auditability of agent-driven orchestration, as well as the governance mechanisms needed to prevent violations of organisational policy or regulatory constraints.

The ecosystem-governance lens addresses the emergent properties of large-scale, heterogeneous multi-agent ecosystems. The central challenge is managing agent populations: discovery, trust calibration, capability assessment, behavioural monitoring, policy enforcement, and collective alignment with human values and societal norms. This lens draws on SOA governance traditions~\cite{ref10}, agent-based software engineering~\cite{ref23}, and emerging AI safety and alignment work.

\subsection{What ASOC Is and Is Not}

The novelty of ASOC requires clear boundary-setting with adjacent research areas. 
Table~\ref{table2} precisely positions ASOC, showing how each adjacent research area differs from it in its primary concern and engineering focus. This boundary-setting is essential for establishing ASOC as a distinct discipline rather than a relabelling of existing fields.

\begin{table*}[!t]
\centering
\caption{Positioning ASOC Relative to Adjacent Research Areas}
\label{table2}
\renewcommand{\arraystretch}{1.25}
\setlength{\tabcolsep}{6pt}

\begin{tabularx}{\textwidth}{
|>{\raggedright\arraybackslash}p{0.23\textwidth}
|>{\raggedright\arraybackslash}p{0.22\textwidth}
|>{\raggedright\arraybackslash}X|
}
\hline
\rowcolor[HTML]{D9EAF7}
\textbf{Adjacent Area} &
\textbf{Primary Concern} &
\textbf{Why ASOC is Different} \\
\hline

Multi-agent systems &
Coordination among autonomous agents &
ASOC focuses on service-engineered, deployable, governable ecosystems with enterprise-grade contracts and lifecycle management \\
\hline

LLM agents/frameworks &
Reasoning, planning, tool use &
ASOC addresses service contracts, lifecycle, QoS, observability, and security: the engineering layer above cognition \\
\hline

AI-as-a-Service &
Serving AI models through APIs &
ASOC concerns autonomous service behaviour and agent-driven orchestration, not only model serving \\
\hline

SOA / microservices &
Interoperable service components &
ASOC extends service principles to probabilistic, goal-directed, stateful, adaptive agents requiring richer contracts \\
\hline

Workflow / BPM &
Process modelling and execution &
ASOC enables goal-driven orchestration where plans are generated, adapted, and governed at runtime \\
\hline

\end{tabularx}
\end{table*}

ASOC must also be distinguished from the emerging body of work on Agentic Service-Oriented Computing as a lifecycle-driven paradigm~\cite{ref17}. That work establishes the lifecycle-oriented paradigmatic foundation: a four-phase lifecycle of design, deployment, operation, and evolution. This manifesto articulates the normative service-engineering doctrine, reference architecture, and community agenda required to consolidate ASOC as a research field. The two contributions are complementary; together, they provide the structural and normative foundations the field requires.

\subsection{Continuity and Departure}

ASOC is positioned as the next stage of service-oriented computing. The fundamental engineering principles of SOC (loose coupling, composability, discoverability, lifecycle management, QoS accountability, and governed interoperability) remain fully operative in ASOC. What changes is the nature of the service entity: from a deterministic, stateless function to a probabilistic, stateful, goal-directed agent.

Two properties distinguish agentic services from conventional services and demand fundamentally new engineering approaches. The first is goal-directedness: an agent is delegated a goal and must determine how to achieve it, introducing a new layer of indirection with significant implications for specification, verification, and accountability. The second is emergence: system-level behaviour arises from agent interactions in ways that cannot be fully predicted from individual agent specifications. Managing emergence, ensuring it produces desirable rather than harmful behaviours, is a new and critical engineering challenge for which SOC provides important but insufficient foundations.

\section{The Principles of Agentic Service-Oriented Computing}

We articulate six foundational engineering principles that define what it means to build agentic systems well. Each principle is stated, its importance explained, core research questions identified, and concrete mechanisms for realisation enumerated.

\subsection{[Principle 1] Harnessability: Agents Must Be Controllable, Auditable, and Revocable}

\emph{Definition:} An agentic service must remain under meaningful human control at all times through: 

\begin{enumerate}[label=\roman*.]    
\item Controllability: The ability to constrain, redirect, or halt agent behaviour without requiring full knowledge of internal agent state; 
\item Auditability: Production of execution traces supporting post-hoc explanation;
\item Revocability: The ability to reverse agent-initiated actions where technically feasible.
\end{enumerate}

\emph{Why it matters:} Harnessability is not a restriction on agent capability; it is the engineering condition that makes capable agents deployable. It is the primary realisation mechanism.

\emph{Core research questions:} How should delegation semantics be formalised, specifying scope, duration, revocability, and accountability conditions? What formal models of controllability are appropriate for probabilistic agents? How can execution traces be structured for real-time monitoring and audit?

\emph{Realisation mechanisms:} Agent harness, harness-enforced approval gates, revocation tokens, policy enforcement points, sandboxed tool calls, circuit breakers.

\subsection{[Principle 2] Composability and Interoperability: Agentic Services Must Interoperate Through Explicit Contracts and Open Standards}

\emph{Definition:} Agentic services must expose precisely described capability interfaces supporting automated discovery, selection, and assembly. Capability contracts specify functional capabilities, goal types, uncertainty characteristics, execution constraints, and resource/trust requirements.

\emph{Why it matters:} Without composability, multi-agent systems remain bespoke assemblies. Without interoperability standards, the field will fragment around proprietary protocols. Emerging standards such as MCP~\cite{ref24} and A2A~\cite{ref29} are important early steps; ASOC must provide the broader engineering framework (capability contracts, lifecycle semantics, QoS models, policy enforcement) within which such protocols can be made dependable.

%\vspace{6pt}
\emph{Core research questions:} What capability description languages are appropriate? How should semantic interoperability be validated? How should ASOC build on and extend OpenAPI\footnote{https://www.openapis.org/}, AsyncAPI\footnote{https://www.asyncapi.com/en}, CloudEvents\footnote{https://cloudevents.io/}, OAuth/OIDC\footnote{https://openid.net/}, and W3C PROV\footnote{https://www.w3.org/TR/prov-overview/}? What governance is needed for agent registries and marketplaces?

%\vspace{6pt}
\emph{Realisation mechanisms:} Capability description language, delegation contract, agent registry and marketplace, MCP/A2A adapters, signed tool manifests, capability verification tests.

\subsection{[Principle 3] Lifecycle Engineering: Agents Must Be Developed, Deployed, and Evolved Rigorously}

\emph{Definition:} An agentic service has a lifecycle from requirements specification through design, implementation, testing, deployment, operation, monitoring, and retirement, which must be subject to formal engineering discipline.

%\vspace{6pt}
\emph{Why it matters:} Unlike conventional services, agentic services exhibit probabilistic, context-dependent behaviour influenced by underlying models. The Agent Development Lifecycle (ADLC) must address goal specification and alignment verification, capability benchmarking, behavioural drift detection, model version management, and management of emergent multi-agent interactions. The SODDM tradition~\cite{ref20} provides the closest precedent.

%\vspace{6pt}
\emph{Core research questions:} How should the ADLC be formally structured? What testing and certification methodologies are appropriate for probabilistic agents? How can behavioural drift be detected and managed? How should agent versioning, rollback, and retirement be governed?

%\vspace{6pt}
\emph{Realisation mechanisms:} ADLC framework, agent regression test suites, behavioural drift monitors, versioned prompt/tool/model registries.

\subsection{[Principle 4] Trustworthiness by Design: Policy, Safety, Privacy, and Fairness Must Be Engineered into Agentic Services}

\emph{Definition:} Trust in agentic systems must be engineered into design from the outset, across five dimensions: factual reliability (accurate, grounded outputs); behavioural alignment (consistent with specified values and constraints); security (resistance to adversarial manipulation); privacy (compliance with regulatory data frameworks); and fairness (no systematic discrimination against protected or vulnerable groups).

%\vspace{6pt}
\emph{Why it matters:} Governance bolted on after deployment cannot achieve the depth of trust that consequential enterprise and societal deployment requires. Each trustworthiness dimension requires specific engineering instruments that must be developed as first-class ASOC research contributions.

%\vspace{6pt}
\emph{Core research questions:} How should trustworthiness properties be formally specified and verified? How can policy compliance be enforced without impairing autonomy? How can fairness properties be monitored at scale?

%\vspace{6pt}
\emph{Realisation mechanisms:} policy enforcement framework, alignment verification suites, privacy guard layers, fairness monitors, regulatory compliance evidence schemas.

\subsection{[Principle 5] Goal-Driven Orchestration: Services Must Be Assembled Around Outcomes}

\emph{Definition:} The shift from procedure-driven to goal-driven orchestration is the defining architectural innovation of agentic services. The orchestrating agent receives a goal specification and determines, at runtime, the service composition strategy needed to achieve it.

%\vspace{6pt}
\emph{Why it matters:} Goal-driven orchestration enables a qualitatively new level of flexibility and adaptability appropriate for tasks whose procedural realisation cannot be fully anticipated at design time. It does not replace procedure-driven approaches; it provides a higher-level abstraction layer.

%\vspace{6pt}
\emph{Core research questions:} What goal specification languages and semantics are needed? How can automated goal decomposition be performed reliably and verifiably? How can goal-directed orchestration be verified against safety and compliance constraints?

%\vspace{6pt}
\emph{Realisation mechanisms:} Goal specification language, automated planning and decomposition engine, runtime constraint solver, goal-progress monitor, exception/recovery handlers.

\subsection{[Principle 6] Observability and Accountability: What Agents Do Must Be Explainable and Auditable}

\emph{Definition:} In any system where agents take consequential actions, those actions must be fully observable and accountable through logged, structured, accessible execution traces spanning the full goal-to-action chain.

%\vspace{6pt}
\emph{Why it matters:} Accountability requires that for every agent action, there exists a clear causal chain from human delegation through agent planning and decisions to action execution, traceable by responsible parties and regulators. An agent provenance graph represents the complete causal structure of an agent's execution: user goals, planning steps, decision points, tool calls, data sources, service invocations, applied policies, approvals, and final actions. Developing, standardising, and querying such graphs are central ASOC research challenges.

%\vspace{6pt}
\emph{Core research questions:} What abstractions (structured execution traces, agent provenance graphs, semantic audit logs) make agentic behaviour accountable without requiring full model interpretability? How should provenance be maintained across multi-agent workflows? How can audit evidence be generated for regulatory reporting?

%\vspace{6pt}
\emph{Realisation mechanisms:} Semantic audit logs, agent provenance graphs, real-time event streams, policy-decision logs, compliance evidence exporters.

\section{The Agentic Services Research Agenda}

The six principles are translated into a five-dimensional research agenda. These dimensions parallel and extend the four themes of the SOC Research Roadmap~\cite{ref8} (foundations, composition, management and monitoring, and engineering), which are reconstituted for the agentic context, with a fifth dimension for evaluation added. A field becomes mature when it has not only engineering methods but also evaluation frameworks, benchmarks, and certification practices; ASOC must build these from the outset.

\subsection{Agentic Service Foundations and Lifecycle Engineering}

This dimension addresses the foundational infrastructure for engineering agentic services as first-class, harnessable entities. It corresponds to the service foundations theme of the SOC Roadmap~\cite{ref8}, extended for LLM-powered agents. The SODDM~\cite{ref20} provides the most mature precedent: a phased lifecycle that the ADLC must extend to accommodate probabilistic, goal-directed agents.

%\vspace{6pt}
\emph{Central research questions:} How should agentic capability be formally described to support automated discovery, selection, and substitution? What is the appropriate semantics for a capability contract, and how can such contracts be verified? How should the ADLC be structured, and what testing, validation, and certification methodologies are appropriate? How can behavioural drift be detected and managed? How should agent versioning, rollback, and retirement be governed?

The concept of an agentic service registry and marketplace also belongs in this dimension. If agents are services, there will be registries, discovery mechanisms, reputation systems, billing models, and certification requirements, extending the publish-find-bind heritage of SOC~\cite{ref1} into the agentic context. Formalising the semantics of delegation (the scope, duration, revocability, and accountability-transfer conditions of human-to-agent delegation) is a foundational challenge for ASOC. Agentic service marketplaces will also require responsibility-allocation mechanisms that clarify provider obligations, consumer obligations, liability boundaries, and certification status.

The accountability structures of agentic service deployment present a genuinely novel legal and engineering challenge. When an agentic service causes harm, responsibility is distributed across a chain of principals that may include the delegating user or organisation, the agent provider, the underlying model provider, the tool or plugin provider, the orchestrating platform, and the deploying organisation. Existing liability frameworks, drawn from product liability, professional negligence, and software service agreements, were designed for systems with clear, traceable causal chains from developer to artefact to harm. Agentic services break these assumptions: harm may arise from an emergent interaction between a model capability, a tool behaviour, a delegated goal, and an environmental condition, none of which, in isolation, was defective or unauthorised. ASOC must therefore treat responsibility allocation as a first-class engineering and governance concern, not a legal afterthought. The delegation contract is a partial mechanism: by specifying the scope, authority, constraints, and accountability conditions of a delegation, it creates an explicit record of what the agent was authorised to do and under what constraints. However, a delegation contract between a user and an agent provider does not by itself resolve the responsibility boundaries between the agent provider, model provider, tool provider, and orchestrating platform. ASOC research must develop formal responsibility-allocation models embedded in capability contracts, certification schemas, and marketplace governance frameworks, clarifying how liability is partitioned, how it transfers between principals as a delegated goal propagates through a multi-agent pipeline, and what evidence an agent provenance graph must preserve to support post-incident accountability in regulated domains.

\subsection{Composition, Orchestration, and Interoperability}

This dimension addresses the design of composable agentic services and the multi-agent systems built from them. It extends the service composition theme of the SOC Roadmap~\cite{ref8}. Existing composition surveys~\cite{ref13,ref28} map the space from manual to automated approaches.

The agentic context introduces a new category in which LLM reasoning can complement manual specification and formal planning by dynamically proposing, adapting, and executing service compositions at runtime. This formulation is important: formal planning and verification remain essential for safety-critical paths, constraint satisfaction, and compliance-bounded orchestration; LLM reasoning extends the compositional space to tasks with full procedural. % specification is impractical.

%\vspace{6pt}
\emph{Central research questions:} What composition models are appropriate for agentic services, and how do they relate to SOA workflow, choreography, and orchestration models? How can automated goal decomposition be performed reliably and verifiably? How should agentic service interfaces and protocols be standardised? ASOC must provide the broader engineering framework within which protocols such as MCP and A2A can be made dependable, extending them with capability contracts, lifecycle semantics, QoS models, and policy enforcement. Existing foundations (OpenAPI, AsyncAPI, CloudEvents, OAuth/OIDC, W3C PROV) offer valuable building blocks.
Table~\ref{tab:standards-gaps} maps each existing artefact to its ASOC contribution and the gaps that remain.

%%%%%%%%%%%%%%%%%%%%%%%%%%%%%%%%%
% Same preamble packages required:
% \usepackage{xcolor}
% \usepackage{colortbl}

\begin{table}[t]
\centering
\caption{Existing Standards and Protocols: Contributions
         to ASOC and Remaining Gaps. Each artefact
         addresses part of the ASOC engineering problem
         but leaves critical concerns unresolved.}
\label{tab:standards-gaps}
\footnotesize
\begin{tabular}{|p{1.5cm}|p{2.1cm}|p{3.9cm}|}
\hline
\rowcolor[HTML]{D9EAF7}
\textbf{Existing Artefact} & \textbf{Helps With} & \textbf{Still Missing for ASOC} \\
\hline
MCP
  & Tool and data access
  & Delegation semantics, Agentic QoS,
    compliance evidence \\
\hline
A2A
  & Agent communication
  & Trust calibration, lifecycle
    management, revocation \\
\hline
OpenAPI
  & Service interface description
  & Goal-level capability contracts,
    uncertainty characteristics \\
\hline
AsyncAPI
  & Async and event-driven service description
  & Goal-event semantics, agent
    subscription contracts \\
\hline
CloudEvents
  & Event interoperability
  & Agent action event schemas,
    provenance binding \\
\hline
OAuth / OIDC
  & Identity and authorisation
  & Delegation-scoped credentials,
    agent identity, revocation \\
\hline
W3C PROV
  & Provenance modelling
  & Agent reasoning and action
    provenance graph \\
\hline
\end{tabular}
\end{table}
%%%%%%%%%%%%%%%%%%%%%%%%%%%%%%%%%%%%

Work on LLM-augmented business process management, ProcessGPT~\cite{procesGPT}, provides an important early bridge between classical process-driven composition and the agent-assisted, goal-directed orchestration that ASOC now systematises.
QoS-aware composition~\cite{ref7,ref13} takes on new complexity in the agentic context. Agent QoS includes not only conventional non-functional properties (latency, throughput, cost) but also agent-specific properties: factual accuracy, reasoning depth, goal-alignment fidelity, and output uncertainty. Extending QoS-aware composition to handle these properties constitutes a major open research problem. The problem of semantic interoperability~\cite{ref6} also resurfaces: LLM-based agents can interpret natural-language capability descriptions and infer semantic compatibility, but this matching is probabilistic and unverified. Research must investigate hybrid approaches combining LLM expressiveness with formal verification rigour.

\subsection{Governance, Observability, and Accountability}

This dimension addresses the management of multi-agent systems at the ecosystem level. It extends the service management and monitoring theme of the SOC Roadmap~\cite{ref8}, drawing on prior work on runtime protocol discovery from service interaction logs~\cite{ref27} as a methodological foundation for behavioural analysis of agentic workflows.
%
%\vspace{6pt}
\emph{Governance must operate at multiple levels:} individual agent, agent-service interaction, workflow, organisation, and ecosystem. 

%\vspace{6pt}
\emph{Central research questions:} What governance architectures are appropriate for multi-agent ecosystems, and how do they manage the tension between agent autonomy and organisational control? How should agent registries and capability directories support dynamic discovery and trust assessment? What monitoring frameworks are needed at the scale of large multi-agent deployments? How should SLAs be defined, negotiated, and enforced between agent principals, providers, and the services they invoke?

A critical governance challenge is managing emergent multi-agent behaviour. System-level properties (collective intelligence, resource contention, failure propagation) emerge from local agent interactions in ways that cannot be directly specified or controlled. ASOC research must develop frameworks to detect harmful emergent behaviours (unintended agent coalitions, resource monopolisation, coordinated misinformation) and to intervene at the governance layer. The alignment of agent ecosystems with regulatory frameworks is a genuinely transdisciplinary challenge that requires methods to translate regulatory requirements into formal agent constraints, verify compliance, and generate audit evidence from execution traces. Human-agent interaction (goal clarification, consent, intervention, explanation, appeal, and shared control) also demands dedicated research as the full interaction spectrum that societal trust requires.

\subsection{Security, Trust, and Risk Management}

Security and trust in agentic systems present qualitatively different challenges from those in conventional service systems. Agentic systems introduce a new attack surface: the natural language interface through which agents receive instructions and tool descriptions. OWASP's LLM Top 10~\cite{ref30} identifies prompt injection as a primary LLM risk, while NIST's AI Risk Management Framework~\cite{ref31} 
provides governance language applicable to trust/compliance in ASOC.

Prompt injection attacks, embedding adversarial instructions in agent inputs or tool outputs, represent a class of vulnerability for which existing service security frameworks provide only partial coverage and for which no mature, standardised ASOC-level countermeasure yet exists. Addressing this gap is a research priority for the field.

Table~\ref{table3} presents the ASOC-specific threat model, mapping seven distinct threat categories to their agentic manifestations and the required ASOC engineering responses. This threat model is distinctive in that it includes threats operating at the natural-language and delegation layers, which have no direct counterpart in conventional service security frameworks.

\begin{table*}[!t]
\centering
\caption{ASOC-Specific Threat Model}
\label{table3}
\renewcommand{\arraystretch}{1.25}
\setlength{\tabcolsep}{6pt}

\begin{tabularx}{\textwidth}{
|>{\raggedright\arraybackslash}p{0.20\textwidth}
|>{\raggedright\arraybackslash}p{0.31\textwidth}
|>{\raggedright\arraybackslash}X|
}
\hline
\rowcolor[HTML]{D9EAF7}
\textbf{Threat} &
\textbf{ASOC-Specific Manifestation} &
\textbf{Required ASOC Response} \\
\hline

Prompt injection &
Adversarial instructions in user input, documents, webpages, or tool outputs &
Runtime instruction hierarchy, input/output filtering, tool-call verification \\
\hline

Tool poisoning &
Malicious or misleading tool descriptions manipulate agent behaviour &
Signed tool metadata, registry governance, capability verification \\
\hline

Privilege escalation &
Agent gains access beyond its delegated authority &
Least-privilege scoped credentials, delegation contracts, policy-based access control \\
\hline

Data exfiltration &
Agent leaks sensitive data through outputs or unauthorised service calls &
Data-loss prevention, privacy guards, output classification, audit logs \\
\hline

Goal drift &
Agent pursues a related but unauthorised goal due to model or prompt changes &
Goal monitoring, constraint checking, harness enforcement, human escalation \\
\hline

Delegation misuse &
Agent performs an action technically allowed by credentials but outside the user's intended delegation scope &
Delegation contract enforcement, intent confirmation, approval gates, policy-decision logs \\
\hline

Multi-agent cascade failure &
Agents reinforce errors or coordinate harmful behaviour at scale &
Multi-agent monitoring, circuit breakers, governance-level isolation \\
\hline

\end{tabularx}
\end{table*}

%\vspace{6pt}
\emph{Central research questions:} How should zero-trust security principles be extended to agent-to-agent communication and agent-to-service invocation? How should agent credential management be governed to prevent privilege escalation? How can supply-chain security be assured when invoking tools from heterogeneous providers? How should privacy-preserving computation be integrated into agent workflows?

Trust calibration, assigning appropriate levels of trust to agents and agent-produced artefacts, is foundational for ASOC. Trust must be earned dynamically through operational track record, validated through capability benchmarking, and calibrated to the stakes of specific delegation tasks. Research must develop formal trust models, reputation mechanisms, and runtime enforcement that constrain agent authority in proportion to established trust.

\subsection{Evaluation, Certification, and Agentic Quality of Service}

A mature engineering discipline requires not only methods for building systems but methods for evaluating them. This fifth dimension addresses the evaluation, benchmarking, and certification of agentic services.

Agentic QoS should be treated as a first-class ASOC construct, analogous to classical QoS in SOC but extended to probabilistic, goal-directed, multi-turn, and policy-constrained execution. ASOC requires a new theory of Agentic Quality of Service that integrates classical measures (latency, throughput, availability, cost) with agent-specific measures: goal-completion fidelity, factual reliability, uncertainty calibration, policy compliance rate, tool-use safety, provenance completeness, and human-oversight effectiveness. Developing validated metrics for each property and compositional models for aggregating them across multi-agent pipelines~\cite{rezaei2025gwise} is a foundational research challenge.
Table~\ref{tab:agentic-qos} operationalises Agentic QoS by mapping each metric to its category and indicative measurement approach.

%%%%%%%%%%%%%%%%%%%%%%%%%%%%%%%%%%
% Same preamble packages required:
% \usepackage{xcolor}
% \usepackage{colortbl}

\begin{table}[t]
\centering
\caption{Agentic QoS Metrics. Classical measures are
         extended with agent-specific properties that
         capture goal fidelity, trust, and accountability
         requirements absent from conventional QoS models.}
\label{tab:agentic-qos}
\footnotesize
\begin{tabular}{|p{1.9cm}|p{1.3cm}|p{4.5cm}|}
\hline
\rowcolor[HTML]{D9EAF7}
\textbf{Metric} & \textbf{Category} & \textbf{Description and Indicative Measure} \\
\hline
Latency
  & Classical
  & Time from goal submission to response delivery;
    measured in ms/s per task \\
\hline
Availability
  & Classical
  & Operational uptime fraction of the agentic service;
    measured as uptime percentage \\
\hline
Cost
  & Classical
  & Computational and financial cost per goal execution;
    measured in tokens or USD per task \\
\hline
Factual Reliability
  & Agentic
  & Accuracy and groundedness of agent-produced outputs;
    measured by hallucination rate and citation precision \\
\hline
Goal Fidelity
  & Agentic
  & Degree to which outcomes satisfy the delegated goal;
    measured by goal-completion score \\
\hline
Policy Compliance
  & Agentic
  & Rate at which agent actions conform to specified
    constraints; measured by constraint-violation rate \\
\hline
Uncertainty Calibration
  & Agentic
  & Alignment between agent confidence and actual
    accuracy; measured by expected calibration error (ECE) \\
\hline
Provenance Completeness
  & Agentic
  & Coverage of the goal-to-action causal chain in audit
    logs; measured by provenance graph coverage ratio \\
\hline
Human-Oversight Effectiveness
  & Agentic
  & Degree to which review gates intercept harmful
    actions before execution; measured by escalation
    interception rate \\
\hline
\end{tabular}
\end{table}
%%%%%%%%%%%%%%%%%%%%%%%%%%%%%%%%%%%

Benchmarking measures performance under standardised tasks; certification establishes whether an agentic service satisfies domain-specific safety, compliance, security, and accountability requirements for deployment. Both are necessary: benchmarking drives comparative research and signals progress, while certification provides the assurance basis that regulated domains require before deployment.

%\vspace{6pt}
\emph{Central research questions:} How do we benchmark an agentic service across repeated updates to models, prompts, policies, or tools? How do we measure task success, safety, compliance, latency, cost, factuality, and user satisfaction jointly? How do we evaluate the execution of long-horizon goals? How do we certify an agentic service for deployment in regulated domains such as healthcare, finance, or public services? How do we test agentic systems under adversarial and failure conditions? How do we define regression testing after model, prompt, policy, or tool updates?

The Services Computing community's strong tradition of QoS measurement and SLA-based evaluation~\cite{ref7,ref13} provides the most important existing foundation for Agentic QoS. Extending this tradition into the probabilistic, goal-directed, multi-turn context of agentic services is one of the most concrete and impactful contributions ASOC can make to the broader AI engineering ecosystem.

\section{Why the Services Community Must Lead}

The case for the services computing community's leadership of Agentic Service-Oriented Computing rests on three foundations: intellectual authority, engineering relevance, and strategic responsibility.

The intellectual authority of the services computing community derives directly from twenty-five years of sustained, rigorous work on precisely the problems that agentic AI must now solve. The SOC community developed the foundational concepts of service composability, QoS management, lifecycle governance, and trusted interoperability that are now urgently needed in the agentic context. The SOC Research Roadmap of 2008~\cite{ref8} identified precisely the research dimensions (foundations, composition, management, and engineering) that define the ASOC research agenda. This is not a coincidence; it reflects that the fundamental challenges of distributed intelligent systems are, at their core, service-oriented.

The engineering relevance of our community is equally clear. Agentic AI's most significant capability gaps are not in cognition; LLMs already demonstrate impressive reasoning and planning~\cite{ref14,ref19}. 
The gaps are in engineering: in the absence of composable interfaces, QoS accountability, lifecycle discipline~\cite{ref20}, security frameworks, and governance infrastructure. These gaps will not be closed by advances in model capability alone; they require the service engineering mindset that our community has developed and validated. The transition from impressive demo to deployable, trustworthy system (the transition that SOA enabled for web services in the early 2000s) requires exactly the kind of engineering discipline that the services computing community provides.

The strategic responsibility of our community is perhaps the most pressing argument. Agentic AI is evolving at an extraordinary speed, with immense commercial investment and intense competitive pressure that discourages the patient, foundational work needed to build dependable systems. Without authoritative scientific leadership, the field will advance through a succession of incompatible proprietary frameworks, each optimised for capability demonstration rather than engineering soundness, creating deeper governance risk.

The Services Computing community is especially well-positioned to address this challenge because it combines technical rigour with an architecturally comprehensive view of systems, ecosystems, and governance structures. We have the intellectual breadth to address the transdisciplinary nature of ASOC (spanning computer science, systems engineering, AI, security, policy, and social science) and the institutional infrastructure (the IEEE Services Congress\footnote{https://services.conferences.computer.org/}, including the ICWS/CLOUD/EDGE/SSE conferences and IEEE Transactions on Services Computing\footnote{https://www.computer.org/csdl/journal/sc}) to organise and sustain a long-term research programme.

%We make this claim not as a territorial assertion but as a statement of historical responsibility. The Services Computing community has the capability to provide the service-engineering spine for a transdisciplinary field: the service contracts, composition models, lifecycle methods, governance mechanisms, observability frameworks, and QoS principles that allow adjacent advances in AI, NLP, security, policy, and human-computer interaction to become dependable service ecosystems. We invite and need the full breadth of adjacent communities (AI agents, multi-agent systems, NLP, AI safety, security, policy, and regulatory research) to co-constitute this field with us.

We make this claim not as a territorial assertion but as a statement of historical responsibility. The Services Computing community can provide the service-engineering spine this transdisciplinary field requires: service contracts, composition models, lifecycle methods, governance mechanisms, and QoS principles that allow adjacent advances in AI, security, policy, and human-computer interaction to become dependable service ecosystems. We invite the full breadth of adjacent communities to co-constitute this field with us.

\section{Conclusion: The Call}

Agentic Service-Oriented Computing (ASOC) is the service-engineering discipline required to transform autonomous and semi-autonomous AI agents into dependable, composable, governable, observable, secure, and accountable service ecosystems. 
%It provides a necessary answer to the engineering crisis that agentic AI is creating at scale, and it is an answer that the services computing community is especially well-positioned and historically responsible to provide.
%
ASOC will not by itself solve model hallucination, alignment, or social governance. Its role is to provide the engineering structures (contracts, harnesses, lifecycle controls, observability, security profiles, and QoS frameworks) through which advances in models, safety, policy, and human-centred design can be made deployable.
%
%An agent becomes an agentic service only when it is engineered as a \emph{discoverable, composable, governable, observable, and accountable service entity with explicit capability, security, lifecycle, and QoS properties}.
The transition from agent to agentic service is precisely the engineering transformation that ASOC is concerned with.
The question is not whether agentic systems will be built and deployed; they already are. The question is whether they will be built well: with the engineering rigour, governance infrastructure, and trustworthiness properties.% that consequential deployment demands. 

%Our twenty-five-year intellectual heritage (in service composition~\cite{ref3,ref4,ref13,ref28}, QoS management~\cite{ref7}, lifecycle engineering~\cite{ref20}, governance~\cite{ref8,ref10}, security, and observability~\cite{ref27}) provides the most important foundation available for addressing the engineering deficits of agentic AI. The six principles (harnessability, composability, lifecycle engineering, trustworthiness by design, goal-driven orchestration, and observability and accountability) define what it means to build agentic systems well. The five-dimensional research agenda maps the work that remains to be done.

We call upon the services computing research and practitioner community to embrace Agentic Service-Oriented Computing (ASOC) as a central priority: in our research programmes, publication venues, funding proposals, and educational curricula. We call upon our community to engage, with urgency and confidence in our intellectual foundations, in the transdisciplinary conversations needed to shape the governance, standards, and policy frameworks of the agentic AI field. We call specifically for the development of reference architectures, open benchmarks, standardised capability contracts, agent registries, Agentic QoS frameworks, security profiles, and governance standards for Agentic Service-Oriented Computing. And we call upon our community to convene, through symposia such as the \textbf{inaugural} 2026 IEEE Symposium on Agentic Services\footnote{\url{https://services.conferences.computer.org/2026/elementor-955/}} and the \textbf{inaugural}  IEEE International Conference on Agentic Services (ICAgentS) 2027, the forums in which researchers, practitioners, policymakers, and industry leaders can work together to realise the vision of agentic services that are dependable, composable, observable, secure, policy-based, and worthy of the trust that humans and organisations will place in them.

\section*{Acknowledgment}
This research was funded by the Centre for Applied Artificial Intelligence at Macquarie University and supported by the Australian Research Council Discovery Project DP230100233. The authors also acknowledge the use of AI-assisted language tools for editorial refinement during manuscript preparation.

\bibliographystyle{IEEEtran}
\bibliography{bibliography}

\end{document}